\documentclass[10pt,twocolumn,letterpaper]{article}

\usepackage{iccv}
\usepackage{times}
\usepackage{epsfig}
\usepackage{graphicx}
\usepackage{amsmath}
\usepackage{amssymb}
\usepackage{booktabs}
\usepackage{multirow}
\usepackage{bm}

\usepackage[hyphens]{url}
\usepackage[hyphenbreaks]{breakurl}

\usepackage[pagebackref=true,breaklinks=true,letterpaper=true,colorlinks,bookmarks=false]{hyperref}

 \iccvfinalcopy 

\def\iccvPaperID{54} 

\ificcvfinal\fi

\begin{document}

\title{Weight Averaging Improves Knowledge Distillation under Domain Shift}

\author{Valeriy Berezovskiy \\
HSE University\\
{\tt\small vsberezovskiy@edu.hse.ru}
\and
Nikita Morozov\\
HSE University\\
{\tt\small nvmorozov@hse.ru}
}

\maketitle
\ificcvfinal\thispagestyle{empty}\fi

\begin{abstract}
   Knowledge distillation (KD) is a powerful model compression technique broadly used in practical deep learning applications. It is focused on training a small student network to mimic a larger teacher network. While it is widely known that KD can offer an improvement to student generalization in i.i.d setting, its performance under domain shift, i.e. the performance of student networks on data from domains unseen during training, has received little attention in the literature. In this paper we make a step towards bridging the research fields of knowledge distillation and domain generalization. We show that weight averaging techniques proposed in domain generalization literature, such as SWAD and SMA, also improve the performance of knowledge distillation under domain shift. In addition, we propose a simplistic weight averaging strategy that does not require evaluation on validation data during training and show that it performs on par with SWAD and SMA when applied to KD. We name our final distillation approach Weight-Averaged Knowledge Distillation (WAKD).
\end{abstract}

\section{Introduction}

Large scale vision models over the years have shown impressive results in various computer vision tasks~\cite{he2016deep, tan2019efficientnet, kolesnikov2020big, dosovitskiy2020image, liu2021swin}. However, the use of such models in practice is often limited by available computational resources. Knowledge distillation~\cite{hinton2015distilling} is a popular compression technique that can be used to improve both memory and computational time efficiency of large neural networks or neural network ensembles. KD transfers the knowledge from a cumbersome teacher model to a more lightweight student model by training the student to match the teacher behavior on the available data instead of predicting the true labels.

Domain generalization (DG)~\cite{blanchard2011generalizingfs} focuses on another challenge that often arises in machine learning applications: domain shift. How well do our models perform when training and test data distributions differ? An example can be a vision model encountering a weather condition or a time of day unseen during training. In DG benchmarks, models are trained on a number of source domains and their performance is evaluated on a different target domain that is unavailable during training to emulate the domain shift setting.

The focus of this paper lies at the intersection of these two fields. How well does the teacher's ability to generalize to unseen domains is transferred to the student during KD? This question naturally arises in the situation when we have a large model that performs well under domain shift and want to compress it into a smaller model, preserving the ability to generalize to unseen domains. Thus the scenario of our interest is when both the teacher training and the distillation are done on source domains, while the performance metric is calculated on the target domain. Despite the large amount of existing research on KD and DG, this setting has received little attention in the existing literature.

In this work, we apply weight averaging techniques to improve knowledge distillation performance under distribution shift. Averaging model weights was proposed to improve neural network training in SWA~\cite{izmailov2018averaging}, where weights of multiple models from a single training trajectory are averaged to construct a better generalizing model (not to be confused with ensembling, where predictions of multiple models are averaged). A number of SWA modifications were created and successfully applied to domain generalization, such as SWAD~\cite{cha2021swad} and SMA~\cite{arpit2022ensemble}, showing that weight averaging improves out-of-distribution generalization as well. We apply SWAD and SMA to knowledge distillation, showing that they improve KD performance across two domain generalization datasets (PACS~\cite{li2017deeper} and OfficeHome~\cite{venkateswara2017deep}) and two teacher-student neural network architectures (ResNet and ViT). In addition, we introduce a simplistic weight averaging strategy by modifying SMA to include all the network weights up to the end of the training trajectory and show that it performs on par with its counterparts. The advantage of this strategy over SWAD and SMA is that it does not need to calculate the validation set performance of the student to choose the averaging segment, thus reducing the computational cost of the whole procedure. We coin such knowledge distillation approach Weight-Averaged Knowledge Distillation (WAKD), constructing a simple yet powerful baseline for future research of knowledge distillation under domain shift. Our source code is available at \url{https://github.com/vorobeevich/distillation-in-dg}.

\section{Related Work}


\textbf{Domain Generalization:} The problem of domain generalization was formally introduced in~\cite{blanchard2011generalizingfs}. Over the years, a plethora of domain generalization methods have been proposed in the literature, including techniques based on domain alignment~\cite{muandet2013domain, motiian2017unified, li2018domain}, meta learning \cite{li2018learning, balaji2018metareg, zhao2021learning}, self-supervised learning~\cite{carlucci2019domain, wang2020learning, albuquerque2020improving, bucci2021self} and data augmentation~\cite{volpi2019addressing, shi2020towards, xu2020robust}. See ~\cite{zhou2022domain} for a comprehensive survey. An important work in the field~\cite{gulrajani2020search} showed that model selection can be crucial, and when correctly taken into account during comparison of the approaches, empirical risk minimization (ERM~\cite{vapnik1998statistical}) baseline showed strong performance in comparison to DG approaches of that time. The main focus of our work is on approaches based on weight averaging. SWAD~\cite{cha2021swad} argues that finding flat minima leads to better performance of the model under domain shift and proposes a dense weight averaging strategy for this purpose. One drawback of SWAD is the fact that it introduces three additional hyperparameters to the training process that are used to find the segment of weights for averaging. This drawback is answered in~\cite{arpit2022ensemble}, where, in addition to other contributions, authors present SMA: an alternative hyperparameter-free weight averaging strategy. 


\textbf{Knowledge Distillation:} Knowledge distillation was proposed in~\cite{hinton2015distilling} as a model compression technique. The original method was to minimize KL-divergence between the teacher's and the student's predictive distributions, thus training the student to match the predictions of the teacher. A number of variations and modifications of the approach were proposed after the original work~\cite{romero2014fitnets, zagoruyko2016paying, tian2019contrastive, zhang2020task, xu2020feature, zhao2022decoupled}. In contrast to developing new distillation objectives, \cite{beyer2022knowledge} showed that the original approach remains a powerful and competitive method in computer vision tasks when taking several design choices into account, such as longer training and passing the same augmented images to the teacher and the student. 

There are several works that employ ideas from KD literature to construct novel methods for domain generalization~\cite{wang2021embracing, lee2022cross, sultana2022self, singh2023robustrl}, however these works do not use knowledge distillation itself as a model compression technique. The authors of~\cite{fang2021mosaicking} consider a setting that resembles ours, with the difference that the goal of the student is to match the teacher's performance on the domain where the teacher was trained, without access to any data from this domain. Among the only works that study the same setting as we do, \cite{zhou2022device} shows that data augmentation techniques such as CutMix~\cite{yun2019cutmix} and Mixup~\cite{zhang2017mixup} improve KD performance under domain shift. The authors of \cite{ojha2022knowledge} observe high agreement between the predictions of the teacher and the student on domains unseen by both of them, which can indicate that KD does in fact transfer knowledge about unseen domains. In another related line of work, \cite{li2023distilling} studies out-of-distribution generalization when distilling vision-language models (e.g. CLIP~\cite{radford2021learning}).



\section{Methodology}

\subsection{Domain Generalization}

The formal setting of domain generalization is as follows. Let $\mathcal{X}$ be the input space and $\mathcal{Y}$ the target space. We have access to data from $K$ source domains $S_k$, each associated with a joint distribution $P_{X Y}^{k}$ on pairs from $\mathcal{X} \times \mathcal{Y}$. The goal of DG is to learn a predictive model $f: \mathcal{X} \rightarrow \mathcal{Y}$ using only the source domains data such that the prediction error on an unseen target domain $T$ is minimized. The corresponding target domain joint distribution is denoted as $P_{X Y}^{T}$, and it is assumed that $P_{X Y}^{T} \neq P_{X Y}^{k}, \forall k \in\{1, \ldots, K\}$.

DG datasets generally contain data from multiple labeled domains. To evaluate the performance of a training approach, all possible splits are considered where one domain is chosen as target and the other ones are chosen as source.

It is important to note that the standard DG setting forbids any use of target domain data during both training and model selection, meaning that both training and validation data come from source domains. However, \cite{gulrajani2020search} showed that picking a model with the best source validation data performance is a viable strategy for model selection in DG.

\subsection{Knowledge Distillation}

In our experiments, we consider image classification problems and employ the broadly used and well-studied distillation objective proposed in the original work~\cite{hinton2015distilling}: KL-divergence between the teacher's and the student's predicted class probability vectors. 

Let $\mathbf{z}_t$ and $\mathbf{z}_s$ be the output class logits predicted by the teacher and the student networks respectively on the same data point. Then the distillation loss is
\begin{equation}
    \mathcal{L}_{\mathrm{KD}}\left(\mathbf{z}_s, \mathbf{z}_t\right):=-\tau^2 \sum_{j=1}^C \sigma_j\left(\frac{\mathbf{z}_t}{\tau}\right) \log \sigma_j\left(\frac{\mathbf{z}_s}{\tau}\right),
\label{kdloss}
\end{equation}
where $\sigma_i(\mathbf{z}):=\exp \left(z_i\right) / \sum_j \exp \left(z_j\right)$ is the softmax function and $C$ is the number of classes. The temperature parameter $\tau$ is introduced to adjust the entropy of the predicted softmax-probability distributions before they are used in the loss computation.  The presented loss does not contain the entropy of the teacher since it does not depend on the predictions of the student.

During distillation, the weights of the teacher network are frozen, while the student network minimizes $\mathcal{L}_{\mathrm{KD}}$, thus learning to match the outputs of the teacher rather than predicting true hard labels. The whole pipeline in our study consists of training the teacher network on the set of available source domains and then distilling it into the student network on the same data. The metric of our interest is the accuracy of the student network on the target domain. To be more precise, it is the improvement of the target domain accuracy when using the distillation objective in comparison to training the same network to predict hard labels.





\subsection{Weight Averaging Strategies}

Weight averaging approaches follow the idea of picking a subset of models from a training trajectory instead of a single model and averaging their weights to create a final better generalizing model.

\textbf{SWAD} evaluates the current model during training with some frequency, picking a segment of training iterations for weight averaging based on validation loss values of the evaluated models. The start of the segment is chosen to be the first iteration where the loss value is no longer decreased during $N_s$ iterations.  The end of the segment is chosen to be the first iteration where the loss value exceeds the loss at the beginning of the segment time tolerance $r$ during $N_e$ iterations. Then, all models for every training iteration inside the segment are averaged. $N_s$, $N_e$, and $r$ are hyperparameters of the method. We refer the reader to~\cite{cha2021swad} for a more detailed description of the algorithm. 

\textbf{SMA} offers a simpler alternative to SWAD. It also averages all models inside some segment of training iterations, but always takes the start of the segment to be some fixed iteration close to the beginning of training. Validation accuracy is computed for models already averaged from the beginning of the segment up to the current iteration instead of individual models. The end of the segment is taken to maximize the validation accuracy of the averaged model.

In the next section, we perform knowledge distillation with SWAD and SMA, 
showing that they produce students with better target domain accuracy in comparison to simply picking an individual model with the best validation performance from a training trajectory. 

Observing that SWAD with distillation loss often behaves in such a way that the end of the segment is chosen to be close to the end of training, we offer an even simpler strategy in comparison to both. We start averaging from some fixed iteration close to the beginning of training (10\% in our experiments) akin to SMA, but average all the weights up to the end of a training trajectory. The advantage of this strategy over SWAD and SMA is that it does not require any model validation during training. \textbf{Thus our proposed WAKD procedure consists of running a regular knowledge distillation and then averaging weights of all models from the training trajectory except for the first 10\%.}



\section{Experiments}

\setlength{\tabcolsep}{5pt}
\begin{table*}[t]
\centering
\begin{center}
\begin{footnotesize}
\begin{sc}
\begin{tabular}{@{}l|ccccc|ccccc@{}}
    \toprule
      &
      \multicolumn{5}{c}{PACS} & \multicolumn{5}{c}{OfficeHome}    \\
     \cmidrule(l){2-11}  
     Model
               & A & C
               & P & S & Avg.
               & A & C
               & P & R & Avg.\\
     \midrule
    ResNet-50 (SWAD)
    & $91.6\!\pm\!0.8$
    & $85.1\!\pm\!1.0$
    & $98.5\!\pm\!0.2$
    & $82.5\!\pm\!0.4$
    & $89.4$
    & $70.8\!\pm\!0.6$
    & $57.5\!\pm\!1.1$
    & $81.3\!\pm\!0.3$
    & $82.5\!\pm\!0.2$
    & $73.0$\\
    \cmidrule(l){1-11} 
    ResNet-18 (SWAD)
    & $82.9\!\pm\!0.2$
    & $76.7\!\pm\!1.9$
    & $94.9\!\pm\!0.2$
    & $77.5\!\pm\!1.6$
    & $83.0$
    & $54.2\!\pm\!0.3$
    & $50.6\!\pm\!1.0$
    & $70.6\!\pm\!0.6$
    & $72.2\!\pm\!0.3$
    & $61.9$\\
    \cmidrule(l){1-11} 
    ResNet-18 KD (ERM)
    & $85.8\!\pm\!0.8$
    & $81.3\!\pm\!1.3$
    & $94.6\!\pm\!0.4$
    & $82.8\!\pm\!0.3$
    & $85.9$
    & $57.8\!\pm\!0.4$
    & $54.6\!\pm\!0.5$
    & $74.4\!\pm\!0.4$
    & $76.2\!\pm\!0.2$
    & $65.8$\\
    ResNet-18 KD (SWAD)
    & $87.4\!\pm\!0.7$
    & $81.9\!\pm\!0.9$
    & $95.2\!\pm\!0.1$
    & $82.1\!\pm\!0.4$
    & $86.7$
    & $59.7\!\pm\!0.1$
    & $55.1\!\pm\!0.8$
    & $74.9\!\pm\!0.4$
    & $76.8\!\pm\!0.3$
    & $66.6$\\
    ResNet-18 KD (SMA)
    & $86.8\!\pm\!0.7$
    & $81.6\!\pm\!1.1$
    & $95.1\!\pm\!0.1$
    & $81.2\!\pm\!0.9$
    & $86.2$
    & $59.7\!\pm\!0.1$
    & $55.0\!\pm\!0.9$
    & $74.9\!\pm\!0.4$
    & $76.9\!\pm\!0.3$
    & $66.6$\\
    ResNet-18 KD (Ours)
    & $87.3\!\pm\!0.2$
    & $81.9\!\pm\!0.2$
    & $95.1\!\pm\!0.2$
    & $82.1\!\pm\!0.1$
    & $86.6$
    & $59.8\!\pm\!0.2$
    & $55.2\!\pm\!0.9$
    & $74.9\!\pm\!0.4$
    & $76.9\!\pm\!0.4$
    & $66.7$\\
    \bottomrule
    \end{tabular}
\end{sc}
\end{footnotesize}
\end{center}
\caption{Knowledge distillation with ResNet models. The teacher is ResNet-50 and the students are ResNet-18. The presented metric is target domain accuracy. 
ResNet-50 shows better performance here in comparison to SWAD paper~\cite{cha2021swad} due to the use of a newer version of pre-trained weights in PyTorch~\cite{paszke2019pytorch}.}
\label{resnets}
\end{table*}

\begin{table*}[t]
\centering
\begin{center}
\begin{footnotesize}
\begin{sc}
\begin{tabular}{@{}l|ccccc|ccccc@{}}
    \toprule
      &
      \multicolumn{5}{c}{PACS} & \multicolumn{5}{c}{OfficeHome}    \\
     \cmidrule(l){2-11}  
     Model
               & A & C
               & P & S & Avg.
               & A & C
               & P & R & Avg.\\
     \midrule
    DeiT-Small (SWAD)
    & $91.9\!\pm\!0.4$
    & $85.1\!\pm\!0.6$
    & $98.9\!\pm\!0.1$
    & $82.4\!\pm\!1.9$
    & $89.6$
    & $72.8\!\pm\!0.7$
    & $61.6\!\pm\!1.1$
    & $81.8\!\pm\!0.8$
    & $83.7\!\pm\!0.1$
    & $75.0$\\
    \cmidrule(l){1-11} 
    DeiT-Tiny (SWAD)
    & $85.2\!\pm\!1.3$
    & $79.0\!\pm\!0.8$
    & $96.9\!\pm\!0.2$
    & $79.8\!\pm\!0.6$ 
    & $85.2$
    & $61.8\!\pm\!0.8$
    & $51.6\!\pm\!1.0$
    & $74.8\!\pm\!0.5$
    & $76.6\!\pm\!0.3$
    & $66.2$\\
    \cmidrule(l){1-11} 
    DeiT-Tiny KD (ERM)
    & $85.8\!\pm\!0.4$
    & $81.2\!\pm\!1.6$
    & $95.5\!\pm\!0.3$
    & $80.1\!\pm\!1.3$
    & $85.7$
    & $63.2\!\pm\!0.5$
    & $56.2\!\pm\!0.9$
    & $78.1\!\pm\!0.6$
    & $79.3\!\pm\!0.5$
    & $69.2$\\
    DeiT-Tiny KD (SWAD)
    & $88.2\!\pm\!0.3$
    & $82.2\!\pm\!0.3$
    & $97.0\!\pm\!0.2$
    & $83.1\!\pm\!0.3$
    & $87.6$
    & $66.4\!\pm\!0.4$
    & $56.6\!\pm\!0.6$
    & $79.0\!\pm\!0.5$
    & $80.2\!\pm\!0.2$
    & $70.5$\\
    DeiT-Tiny KD (SMA)
    & $88.2\!\pm\!0.4$
    & $82.2\!\pm\!0.4$
    & $96.8\!\pm\!0.1$
    & $82.3\!\pm\!1.2$
    & $87.4$
    & $65.8\!\pm\!0.1$
    & $56.2\!\pm\!0.2$
    & $78.9\!\pm\!0.8$
    & $80.3\!\pm\!0.1$
    & $70.3$\\
    DeiT-Tiny KD (Ours)
    & $88.0\!\pm\!0.2$
    & $82.3\!\pm\!0.3$
    & $96.9\!\pm\!0.2$
    & $83.1\!\pm\!0.4$
    & $87.6$
    & $66.1\!\pm\!0.4$
    & $56.8\!\pm\!0.7$
    & $78.9\!\pm\!0.6$
    & $80.3\!\pm\!0.1$
    & $70.5$\\
    \bottomrule
    \end{tabular}
\end{sc}
\end{footnotesize}
\end{center}
\caption{Knowledge distillation with DeiT models. The teacher is DeiT-Small and the students are DeiT-Tiny. The presented metric is target domain accuracy.}
\label{deits}
\end{table*}

\subsection{Experimental Details}

We carry out the experimental evaluation on two domain generalization datasets: PACS~\cite{li2017deeper} (9,991 images, 7 classes, and 4 domains) and OfficeHome~\cite{venkateswara2017deep} (15,588 images, 65 classes, and 4 domains).

We explore two teacher-student architectures: ResNet-50 to ResNet-18~\cite{he2016deep} and DeiT-Small to DeiT-Tiny~\cite{touvron2021training}. It is important to note that the authors of DeiT presented their own novel distillation approach~\cite{touvron2021training}, while we use the original KD objective (see Eq.~\ref{kdloss}) to create a unified experimental setup.

There is no particular necessity in using the best possible performing teacher in our experimental setting since the interest lies mainly in the performance of KD, thus we chose SWAD to train the teachers as an approach well-established in domain generalization literature.

We train teacher models using SWAD for 5,000 iterations (the same amount as in SWAD paper), as well as baselines for student models, where a model of student architecture is trained to predict hard labels independently of the teacher using SWAD. Following observations from~\cite{beyer2022knowledge}, we increase the amount to 50,000 iterations for all distillation experiments. For both SWAD and SMA, we use the same validation frequency of 100 iterations for a fair comparison. For SMA and WAKD, we fix the starting iteration for averaging at 5,000 (10\% of the training duration). All models are initialized with ImageNet-1k~\cite{russakovsky2015imagenet} pre-trained weights at the beginning of the training.

In addition to weight-averaged distillation students, we present the performance of a single model from the iteration with the best validation accuracy (denoted as ERM in the results), acting as a baseline for weight averaging strategies.

All models are trained using Adam~\cite{kingma2014adam} with learning rate of 5e-5 and batch size 64. We set temperature parameter $\tau = 5$ for KD. For SWAD we use standard hyperparameter values from the paper: $N_s = 3$, $N_e = 6$ and $r = 1.3$.

We use a standard set of data augmentations described in~\cite{gulrajani2020search}, which is widely utilized in domain generalization literature. The only difference is that in the case of distillation we use more aggressive “inception-style” random crops~\cite{szegedy2015going} as proposed in~\cite{beyer2022knowledge}.

For each experiment, we present mean and standard deviation values computed across three random seeds. For each seed, we split source domains data into 80\% and 20\% train/val splits as described in~\cite{gulrajani2020search}, then train the teacher and distill it into the student. To allow fair comparison, all averaging strategies are compared on the same training trajectories, meaning that for each seed (and for each choice of target domain) both teacher training and distillation are done only once. Overall, running the presented experiments took 75 GPU days on NVIDIA V100.

\subsection{Results}

The results of the experiments are presented in Table~\ref{resnets} and Table~\ref{deits}. The distillation without weight averaging (ERM) outperforms the baseline of independently trained student networks across both datasets and both architectures, indicating that the teacher's ability to generalize out-of-domain is transferred to the student to some extent and KD can perform well under domain shift. On top of that, adding our weight averaging strategy to KD leads to extra target domain accuracy improvement of +0.8pp on average in the case of ResNets and +1.6pp in the case of DeiTs. When compared to the other two strategies, WAKD shows similar performance to distillation with SWAD and slightly outperforms distillation with SMA, while being simpler and lacking the need to compute validation performance during training. However, in all cases there is still a significant gap between the target domain performance of the teacher and the student, which indicates that there is room for improvement.


\section{Conclusion}

In this paper, we studied the setting of performing knowledge distillation under domain shift. We showed that weight averaging techniques from the domain generalization literature, namely SWAD and SMA, can improve the performance of distillation students on unseen domains. We proposed a novel simplified weight averaging approach that does not require evaluation on the validation set during training and showed its effectiveness when applied to KD under domain shift. Better understanding the reasons behind the said improvements and their connection to flat minima, as well as exploring other distillation objectives are some of the possible future work directions.

\section*{Acknowledgements}
The authors are grateful to Dmitry Vetrov for valuable discussions and feedback. 
This research was supported in part through computational resources of HPC facilities at HSE University.

{\small
\bibliographystyle{ieee_fullname}
\bibliography{egbib}
}

\end{document}